\documentclass{article}
\usepackage[utf8]{inputenc}
\usepackage{amsmath}
\usepackage{amssymb}
\usepackage{mathrsfs}
\usepackage{graphicx}
\usepackage{hyperref}
\usepackage{caption}
\usepackage{subcaption}
\usepackage{algorithm}
\usepackage[noend]{algpseudocode}

\title{Accelerated Hierarchical Density Clustering}
\author{Leland McInnes and John Healy}
\date{\today}

\begin{document}

\maketitle

\begin{abstract}
We present an accelerated algorithm for hierarchical density based clustering. Our new algorithm improves upon HDBSCAN*, which itself provided a significant qualitative improvement over the popular DBSCAN algorithm. The accelerated HDBSCAN* algorithm provides comparable performance to DBSCAN, while supporting variable density clusters, and eliminating the need for the difficult to tune distance scale parameter $\epsilon$. This makes accelerated HDBSCAN* the default choice for density based clustering.
\end{abstract}

\section{Introduction}
 
Clustering is the attempt to group data in a way that meets with human intuition.  Unfortunately, our intuitive ideas of what makes a `cluster' are poorly defined and highly context sensitive \cite{Hennig2015clusters}.  This results in a plethora of clustering algorithms each of which matches a slightly different intuitive notion of what a natural grouping is.  

Despite the uncertainty underlying the clustering process it continues to be used in a multitude of scientific domains.  The fundamental problem of finding groupings is pervasive and results, however poor, are still important and informative. It is used in diverse fields such as molecular dynamics \cite{melvin2016uncovering}, airplane flight path analysis \cite{wilson2016exploratory}, crystallography \cite{spackman2016high}, and social analytics \cite{korakakis2016xenia}, among many others.

While clustering has many uses to many people, our particular focus is on clustering for the purpose of exploratory data analysis. By exploratory data analysis we mean the process of looking for ``interesting patterns'' in a data set, primarily with the goal of generating new hypotheses or research questions about the data set in question. This necessitates minimal parameter selection and few apriori assumptions about the data.  In this use case, it is highly desirable that solutions have informative failure modes.  Specifically, when data is poorly clustered or does not contain clusters, it is necessary to have some indication of this from the clustering algorithm itself.

Many traditional clustering algorithms are poorly suited to exploratory data analysis tasks. In particular, most clustering algorithms suffer from the problems of difficult parameter selection, insufficient robustness to noise in the data, and distributional assumptions about the clusters themselves.

Many algorithms require the selection of the number of clusters, either explicitly, or implicitly through proxy parameters. In the majority of use cases we have encountered, selecting the number of clusters is very difficult apriori.  Methods to determine the number of clusters such as the elbow method and silhouette method are often subjective and can be hard to apply in practice.  Ultimately these methods all hinge on the clustering quality measure chosen; these are diverse and often highly related with particular clustering algorithms \cite{Hennig2015clusters}.

Many practitioners fail to distinguish between partitioning and clustering to the point where the terms are now often used interchangeably.  By clustering we specifically mean finding subsets of the data which group ``naturally'', without necessarily assigning a cluster for all points.  Partitioning, on the other hand, requires that every data point be associated with a particular cluster.  In the presence of noise the partitioning approach can be problematic.  Even without noise, if clear clusters are not present, partitioning will simply return a poor solution.  

\begin{figure}[!htb]
    \centering
    \includegraphics[width=0.8\textwidth]{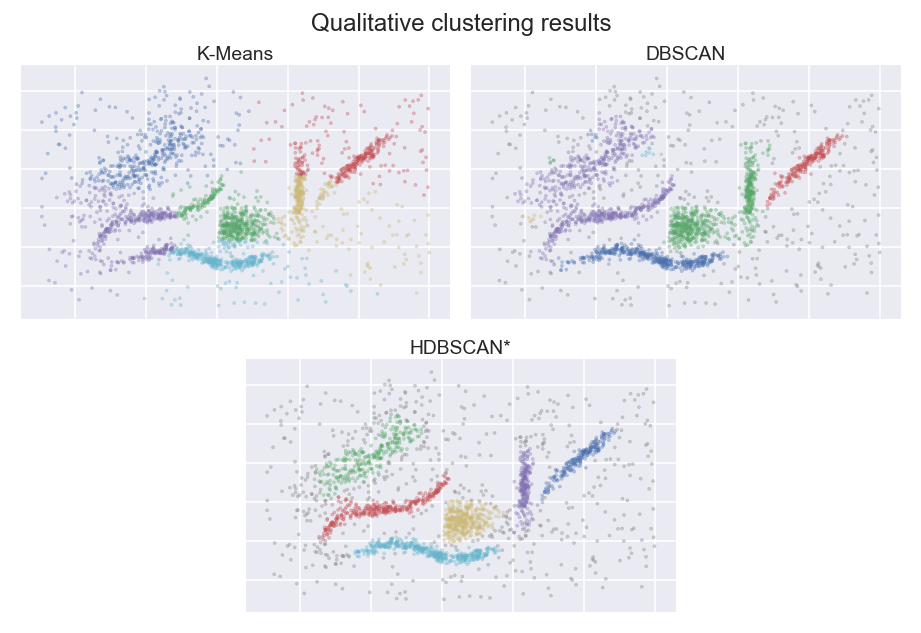}
    \caption{A qualitative comparison of some candidate clustering algorithms on synthetic data. Colors indicate cluster membership (grey denotes noise).  We advocate density based clustering methods when performing exploratory data analysis as they require fewer assumptions about the data distribution, and can refuse to cluster points. Unclustered ``noise'' points for both DBSCAN and HDBSCAN* are depicted in gray.  The above clustering results represent the result of a qualitative hand tuned search for optimal parameters.\protect\footnotemark}
    \label{fig:qualitative_clustering}
\end{figure}

\footnotetext{See \url{https://github.com/lmcinnes/hdbscan_paper/blob/master/Qualitative\%20clustering\%20results.ipynb} for code used to generate these plots}

Distributional assumptions on the data are difficult to make in exploratory data analysis. As a result we examine density based clustering since it has few implicit assumptions about the distribution of clusters within the data.  Among density based clustering techniques DBSCAN \cite{ester1996dbscan} is attractive in that it is efficient and is robust to the presence of noise within data.  Its primary difficulties include parameter selection and the handling of variable density clusters. In \cite{campello2013density} and \cite{campello2015hierarchical} Campello, et al. propose the HDBSCAN* algorithm which addresses both of these problems, but its major difficulty is that it sacrifices performance to do so.

In Figure \ref{fig:qualitative_clustering} we compare three candidate clustering algorithms: K-Means, DBSCAN, and HDBSCAN*. The archetypal clustering algorithm, K-Means, suffers from all three of the problems mentioned previously: requiring the selection of the number of clusters; partitioning the data, and hence assigning noise to clusters; and the implicit assumption that clusters have Gaussian distributions. In comparison, being a density based approach, DBSCAN only suffers from the difficulty of parameter selection. Finally HDBSCAN* resolves many of the difficulties in parameter selection by requiring only a small set of intuitive and fairly robust parameters.

Section \ref{explain} introduces the HDBSCAN* algorithm. We provide three different descriptions of the algorithm: the first description follows Chauduri et al. \cite{Chaudhuri:2010:RCC:2997189.2997228}, \cite{chaudhuri2014consistent} and Stuetzle et al. \cite{stuetzle2003estimating}, \cite{stuetzle2010generalized}, viewing the algorithm as a statistically motivated extension of Single Linkage clustering; the second description follows Campello et al. \cite{campello2013density}, \cite{campello2015hierarchical}, viewing the algorithm as a natural hierarchical extension of the popular DBSCAN algorithm; the third, novel, description is in terms of techniques from topological data analysis \cite{carlsson2009topology}. All three descriptions are valid, and collecting them here serves to bring these diverse fields together. Both the statistical and computational descriptions of HDBSCAN* have been published before (though little comparison has been drawn). We believe the topological description is a significant new contribution of this paper, and offers the opportunity to bring new and powerful mathematical tools to bear on the problem.

The major contribution of this paper is section \ref{accel}, which describes a new algorithm for computing HDBSCAN* clustering results. This new algorithm, building on the work of March et al. \cite{march2010fast} and Curtin et al. \cite{curtin2015faster}, \cite{curtin2013tree}, offers significant improvements in average case asymptotic performance.

In section \ref{perf} we compare the performance of our new HDBSCAN* algorithm against other clustering algorithms. In particular, we demonstrate the asymptotic performance improvement over the reference HDBSCAN* algorithm, and show our new algorithm provides HDBSCAN* with comparable asymptotic performance to DBSCAN, one of the fastest extant clustering algorithms.

\section{HDBSCAN* Explained Three Ways}\label{explain}

Algorithms like HDBSCAN* lie at the convergence of several lines of research from different fields. To highlight this convergence we will describe the HDBSCAN* algorithm from three different perspectives: from a statistically motivated point of view; with a computationally motivated mindset; and in a topologically motivated framework. Through this repetition we hope to both provide a sound introduction to how the algorithm works, and to place it in a richer context of ideas. We also hope that explanations that are less familiar will become easier to follow by analogy to explanations closer to the reader's field of expertise. Finally, we hope to bring together disparate fields of research that are attacking the same problem and arriving at nearly the same solution, and unify their approaches.

\subsection{Statistically Motivated HDBSCAN*}\label{stats}

\begin{figure}[!hbt]
    \centering
    \includegraphics[width=0.5\columnwidth]{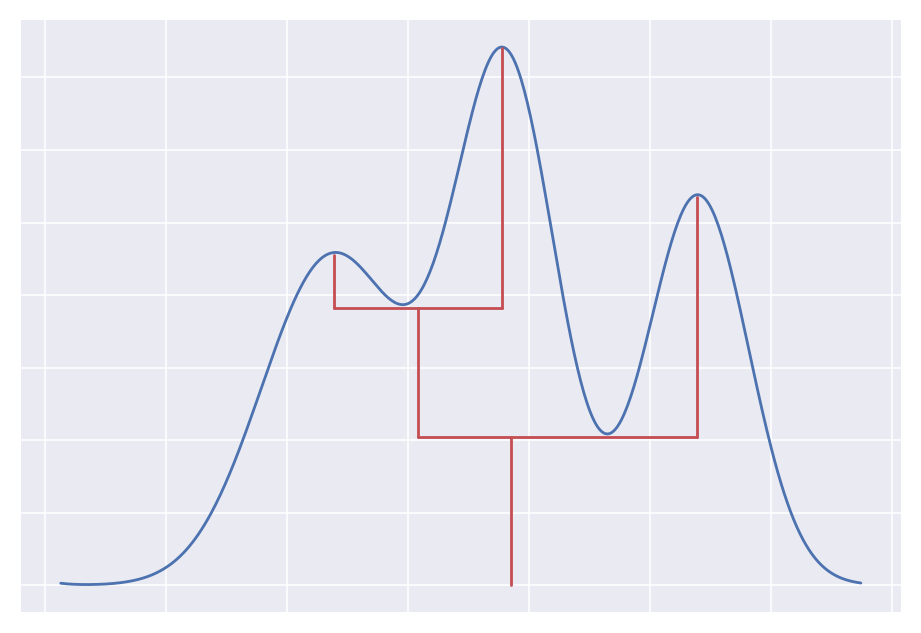}
    \caption{The cluster tree (red) induced by a density function (blue).}
    \label{fig:cluster_tree}
\end{figure}

A statistically oriented view of density clustering begins with the assumption that there exists some unknown density function from which the observed data is drawn. From the density function $f$, defined on a metric space $(\mathcal{X}, d)$, one can construct a hierarchical cluster structure, where a cluster is a connected subset of an $f$-level set $\{x \in (\mathcal{X}, d) \mid f(x) \geq \lambda\}$. As $\lambda\geq 0$ varies these $f$-level sets nest in such a way as to construct an infinite tree, which is referred to as the \emph{cluster tree} (see figure \ref{fig:cluster_tree} for an example). Each cluster is a branch of this tree, extending over the range of $\lambda$ values for which it is distinct. The goal of a clustering algorithm is to suitably approximate the cluster tree, converging to it in the limit of infinite observed data points.

This idea dates back at least to Hartigan \cite{hartigan1981consistency}, and has become an increasingly popular way to frame the clustering problem; see \cite{rinaldo2012stability}, \cite{rinaldo2010generalized}, \cite{stuetzle2003estimating}, \cite{stuetzle2010generalized} and \cite{von2005towards} for examples. Our description of HDBSCAN* in these terms follows Chaudhuri et al. \cite{Chaudhuri:2010:RCC:2997189.2997228}, \cite{chaudhuri2014consistent} and their description of Robust Single Linkage.

The motivation for the approach is based on Hartigan's work on consistency results for single linkage clustering \cite{hartigan1981consistency}. Hartigan's results while impressive, only apply to one dimensional data. The commonly cited drawback of single linkage clustering is that it is not robust to noise and suffers from chaining effects (spurious points merging clusters prematurely) \cite{martinez2016properties}, \cite{wishart1969mode}. Wishart proposed a heuristic algorithm as a potential solution to this in \cite{wishart1969mode}. The Robust Single Linkage algorithm \cite{Chaudhuri:2010:RCC:2997189.2997228}, \cite{chaudhuri2014consistent} extends Wishart's basic approach, and provides suitable theoretical underpinnings.

The Robust Single Linkage algorithm assumes that the data set 
\[
X = \{X_1, X_2, \ldots, X_N\}
\]
is sampled from an unknown density $f$ on some metric space $(\mathcal{X}, d)$. We then define $B(X_i, \varepsilon)$ to be the open ball of radius $\varepsilon$ in $(\mathcal{X}, d)$. The algorithm takes two inputs, $k$ and $\alpha$.
For each $X_i \in X$ define 
$$r_k(X_i) = \inf \{\varepsilon \mid B(X_i, \varepsilon) \text{ contains }k\text{ points}\}.$$
For each $\varepsilon \geq 0$ define a graph $G_\varepsilon$ with vertices $\{X_i \in X \mid r_k(X_i) \leq \varepsilon\}$ and an edge $(X_i, X_j)$ if $d(X_i, X_j) \leq \alpha\varepsilon$. Define the clusters at level $\varepsilon$ of the tree to be the connected components of $G_\varepsilon$.

In \cite{Chaudhuri:2010:RCC:2997189.2997228} and \cite{chaudhuri2014consistent} Chaudhuri et al. provide a number of results on the consistency and convergence of this algorithm in Euclidean space ($\mathbb{R}^d$) for $k \sim d \log N$ with $\sqrt{2} \leq \alpha \leq 2$. Eldridge et al. \cite{eldridge2015beyond}
provide even stronger consistency results by introducing stricter notions of consistency. This provides a sound statistical basis for the approach.

A remaining issue with this algorithm is that the resulting cluster tree with $N$ leaves is highly complex, making analysis difficult for large data set sizes. This is, of course, an issue faced by many hierarchical clustering algorithms. Several authors, including Stuetzle et al. \cite{stuetzle2003estimating}, \cite{stuetzle2010generalized}, and Chaudhuri et al. \cite{chaudhuri2014consistent} have proposed approaches to pruning the cluster tree to simplify presentation and analysis. While Chaudhuri et al. provide consistency guarantees for their approach, we find the required parameters to be less intuitive, and harder to tune. We therefore will follow the ``runt pruning'' algorithm of Stuetzle \cite{stuetzle2003estimating}.

Tree simplification begins with the introduction of a new parameter $m$, the minimum cluster size. Any branch of the cluster tree that represents a cluster of less than $m$ points is pruned out of the tree, and we record the $\varepsilon$ value of the split, defining it as the $\varepsilon$ value when the points of the pruned branch left the parent branch. That is, for each branch $C_i$ of the cluster tree there is an associated set of points $\{X_{i_1}, X_{i_2}, \ldots, X_{i_t}\} \subseteq X$, and for each point $X_{i_\ell}$ in $\{X_{i_1}, X_{i_2}, \ldots, X_{i_t}\}$ there exists a value $\varepsilon_\ell$ for which the point $X_{i_\ell}$ is deemed to have left the cluster (including because the cluster $C_i$ split, or because it was removed). 

The resulting pruned tree has many fewer branches, and hence fewer leaves. Furthermore, each remaining branch has a record of the points remaining in the branch at each $\varepsilon$ value for which the branch exists. The result is a far simpler tree of clusters, amenable to further analysis, but still containing rich information about the actual cluster structures at a point-wise level.

Finally, it is often desirable to extract a flat clustering -- selecting a set of non-overlapping clusters from the tree. For hierarchical cluster schemes this often takes the form of choosing a ``cut level'' (in our case a choice of $\varepsilon$) and using the clustering at that level of the tree. When we wish to consider variable density clusters, the cut level varies through the tree, and thus we must choose a different approach to selecting a flat clustering.

\begin{figure}[!htb]
    \centering
    \begin{subfigure}[b]{0.48\textwidth}
        \includegraphics[width=\textwidth]{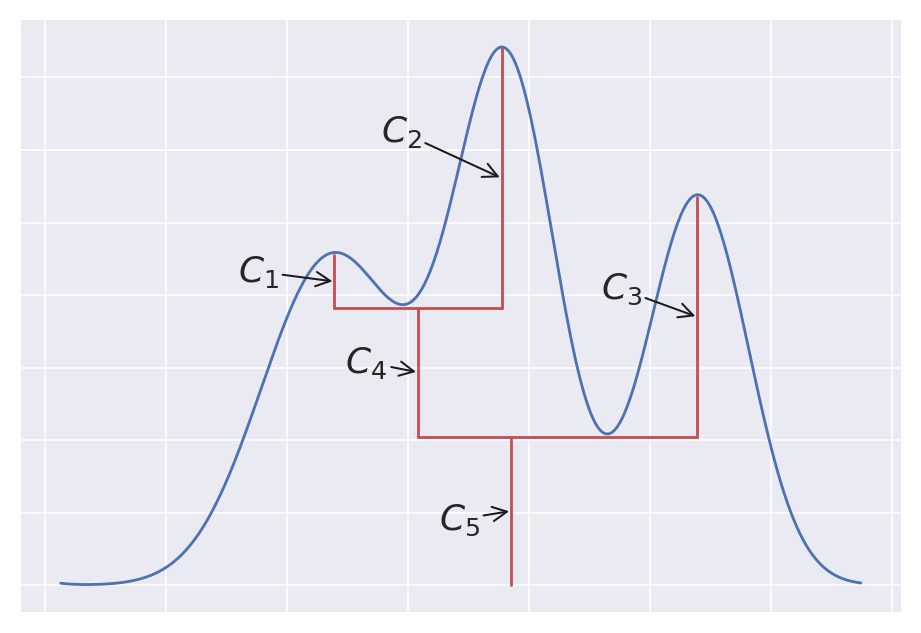}
        \caption{}
    \end{subfigure}
    \begin{subfigure}[b]{0.48\textwidth}
        \includegraphics[width=\textwidth]{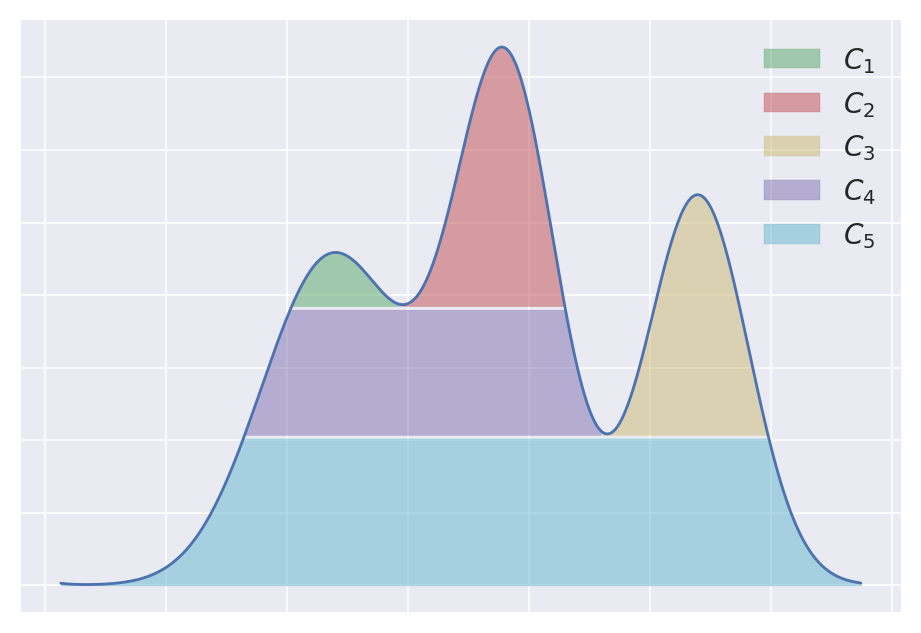}
        \caption{}
    \end{subfigure}
    \caption{Relative excess of mass for the cluster tree from figure \ref{fig:cluster_tree}. In (a) we label the clusters, and (b) depicts the relative excess of mass as shaded areas, using different colours for each clusters relative excess of mass.}
    \label{fig:excess_of_mass}
\end{figure}

Notionally our goal is to determine the clusters that persist over the largest ranges of distance scales. To do this we require a measure of the persistence of a cluster. To make this concrete we refer again to Hartigan \cite{hartigan1987estimation}, and also to M\"uller and Sawitzki \cite{muller1991excess}, for the notion of excess of mass. Given a density function $f$, let $C$ be a subset of the domain of $f$, and define the \emph{excess of mass} of $C$ at a level $\lambda$ to be
\[
E(C, \lambda) = \int_{C_\lambda} (f(x) - \lambda) dx ,
\]
where $C_\lambda = \{x\in C \mid f(x) \geq \lambda\}$. Given a cluster tree for $f$, we can define the excess of mass of a cluster $C_i$ that exists at level $\lambda_{C_i}$ of the cluster tree as follows: Let $\lambda_{\text{min}}(C_i)$ be the minimal $\lambda$ value for the branch associated to $C_i$ in the cluster tree. Then define the excess of mass of $C_i$ to be
\[
E(C_i) = \int_{C_i} (f(x) - \lambda_{\text{min}}(C_i)) dx .
\]
Next we follow \cite{campello2013density} in defining the \emph{relative excess of mass} for a cluster $C_i$. First we define $\lambda_{\text{max}}(C_i)$ to be the maximal lambda value for which $C_i$ exists as a distinct cluster (i.e. before it splits into sub-clusters in the cluster tree). Then the relative excess of mass is
\[
E_R(C_i) = \int_{C_i} (\min(f(x), \lambda_{\text{max}}(C_i)) - \lambda_{\text{min}}(C_i) dx .
\]
Alternatively, if $C_{i_1}, C_{i_2}, \ldots, C_{i_k}$ are the children of $C_i$ in the cluster tree then
\[
E_R(C_i) = E(C_i) - \sum_{j=1}^k E(C_{i_j}) .
\]
That is, the relative excess of mass of a cluster is the total mass of the cluster \emph{not including} the mass of any descendant clusters in the cluster tree. We see this demonstrated in figure \ref{fig:excess_of_mass} with the shaded areas indicating the excess of mass of the each for clusters from the cluster tree.

We can translate these notions to the empirical pruned tree described above. The pruned tree can be used to construct a discrete density function ranging over data points.  In order to do this we require two things.  Firstly, a density associated to each data point.  This is simply the inverse of the $\varepsilon$ value at which the point left the tree (this was the data we recorded in addition to pruning branches of the tree).  Secondly, we need an ordering on the data points such that the cluster tree of the density function is isomorphic to the pruned tree.  This is simply a matter of sorting the points via a depth first search of the pruned tree (making use of the per point $\varepsilon$ values to order data points within a branch of the cluster tree).  Explicitly we have an empirical density 
\[
\hat{f}(X_j) = \frac{1}{\varepsilon_{X_j}}
\]
where $\varepsilon_{X_j}$ is the $\varepsilon$ value at which we recorded the point $X_j$ leaving the tree.  Further we have a cluster tree associated with $\hat{f}$ isomorphic to the pruned tree and we can define $\lambda_{\text{min}}(C_i)$ for any cluster $C_i$ in the pruned tree accordingly.

This allows us to compute excess of mass for $\hat{f}$ and any cluster $C_i$ from the pruned tree as
\[
E(C_i) = \sum_{X_j \in C_i} (\hat{f}(X_j) - \lambda_{\text{min}}(C_i))
\]
and consequently we have a persistence score provided by the relative excess of mass. That is, given cluster $C_i$ having children $C_{i_1}, C_{i_2}, \ldots, C_{i_k}$, we define the persistence score
\[
\sigma(C_i) = E(C_i) - \sum_{j=1}^k E(C_{i_j}) .
\]

The optimal flat clustering can then be described as the solution to a constrained optimization problem. If the set of clusters is $\{C_1,C_2, \ldots , C_n\}$ then we wish to select $I \subseteq \{1, 2, \ldots, n\}$ to maximize
\[
\sum_{i\in I} \sigma(C_i)
\]
subject to the constraint that, for all $i, j \in I$ with $i\neq j$, we have
\[
C_i\cap C_j = \emptyset .
\]
That is, we wish to maximize the total persistence score over chosen clusters, subject to the constraint that clusters must not overlap. This constrained optimization problem is can be solved in a straightforward manner \cite{campello2015hierarchical}.

\subsection{Computationally Motivated HDBSCAN*}\label{comp}

HDBSCAN* can be thought of as a natural extension of the popular DBSCAN algorithm. We begin, following \cite{campello2013density} and \cite{campello2015hierarchical}, by describing a modified version of DBSCAN, denoted DBSCAN*, that will make the relationship clearer. This algorithm is an adaptation of standard DBSCAN which removes the notion of border points. Removing border points provides clarity and improves consistency with the statistical interpretation of clustering in section \ref{stats}. DBSCAN* takes two parameters, $\varepsilon$ and $k$, where $\varepsilon$ is a distance scale, and $k$ is a density threshold expressed in terms of a minimum number of points.  Extending to HDBSCAN* can be conceptually considered as searching over all $\varepsilon$ values for DBSCAN* to find the clusters that persist for many values of $\varepsilon$.  This selection of clusters, which persist over many distance scales, provides the benefits of not only eliminating the need to select the $\varepsilon$ parameter but also of dealing with the problem of variable density clustering, something which classical DBSCAN struggles with.  

We will describe DBSCAN* in the same terms used by Campello, et al. \cite{campello2013density}. Again we will be working with a set of $X = \{X_1, X_2, \ldots, X_N\}$ of data points in a metric space $(\mathcal{X}, d)$.

A point $X_i$ is called a \emph{core point} with respect to $\varepsilon$ and $k$ if its $\varepsilon$-neighbourhood contains at least $k$ many points, i.e. if $|B(X_i,\varepsilon)\cap X| \geq k$.  That is, the open ball of radius $\varepsilon$ contains at at least $k$ many points from X.  

Two core points $X_i$ and $X_j$ are \emph{$\varepsilon$-reachable } with respect to $\varepsilon$ and $k$ if $X_i \in B(X_j,\varepsilon)$ and $X_j \in B(X_i,\varepsilon)$.  That is, they are both core points with respect to $k$, and are both contained within each others $\varepsilon$-neighbourhood. 
Two core points $X_i$ and $X_j$ are \emph{density-connected} with respect to $\varepsilon$ and $k$ if they are directly or transitively $\varepsilon$-reachable.  

A \emph{cluster} $C$, with respect to $\varepsilon$ and $k$, is a non-empty maximal subset of $X$ such that every pair of points in $C$ is density-connected. This definition of cluster results in the DBSCAN* algorithm.  

To extend the algorithm to get HDBSCAN* we need to build a hierarchy of DBSCAN* clusterings for varying $\varepsilon$ values. The key to doing this is to redefine how we measure distance between points in $X$. For a given fixed value $k$, we define a new distance metric derived from the metric $d$, called the \emph{mutual reachability distance}, as follows. For any point $X_i$ we define the \emph{core-distance} of $X_i$, denoted $\kappa(X_i)$ to be the distance to the $k^\text{th}$ nearest neighbor of $X_i$; then, given points $X_i$ and $X_j$ we define
\[
d_{\text{mreach}}(X_i, X_j) = \begin{cases}
\max \{ \kappa(X_i), \kappa(X_j), d(X_j, X_j) \} & X_i \neq X_j\\
0 & X_i = X_j
\end{cases}.
\]
It is straightforward to show that this is indeed a metric on $X$. We can then apply standard Single Linkage Clustering \cite{sibson1973slink} to the discrete metric space $(X, d_{\text{mreach}})$ to obtain a hierarchical clustering of $X$. The clusters at level $\varepsilon$ of this hierarchical clustering are precisely the clusters obtained by DBSCAN* for the parameter choices $k$ and $\varepsilon$; in this sense we have derived a hierarchical DBSCAN* clustering.

The goal of a density based algorithm, such as DBSCAN, is to find areas of the greatest density.  To do this we need to shift from the notion of distance to a notion of density. An efficient estimate of the local density at a point can be provided by the reciprocal of the distance to its $k^\text{th}$ nearest neighbor. This is simply the inverse of the core-distance of that point. With this in mind we will work in terms of varying density instead of varying distance by constructing our cluster tree with respect to $\lambda = \frac{1}{\varepsilon}$ and consider the $\lambda$ value at which cluster splits occur.

The next step in the algorithm is to produce a condensed tree that simplifies the hierarchy. For this we introduce a new parameter $m$ which will denote the minimum cluster size that will be accepted. We process the tree from the root downward. At each cluster split we consider the child clusters. Any child cluster containing fewer than $m$ points is considered a spurious split, and we denote those points as ``falling out of the parent cluster'' at the given $\lambda$ value. If only one child cluster contains more than $m$ points we consider it the continuation of the parent, persisting the parent cluster's label/identity to it. If more than a single child cluster contains more than $m$ points then we consider the split to be a ``true'' split. In this fashion we arrive at a tree with a much smaller number of clusters which ``shrink'' in size as they persist over increasing $\lambda$ values of the tree. One can consider this a form of smoothing of the tree.

We can now define the stability of a cluster to be the sum of the range of $\lambda$ values for points in a cluster. Explicitly, we define $\lambda_{\text{max}, C_i}(X_j)$ to be the $\lambda$ value at which the point $X_j$ falls out of the cluster $C_i$ (either as an individual point, or as a cluster split in the condensed tree). Similarly we define $\lambda_{\text{min}, C_i}(X_j)$ as the  minimum lambda value for which $X_j$ is present in $C_i$. Then the stability of the cluster $C_i$ is defined as

\[
\sigma(C_i) = \sum_{X_j \in C_i} \left(\lambda_{\text{max}, C_i}(X_j) - \lambda_{\text{min}, C_i}(X_j)\right) .
\]

The optimal flat clustering can then be described as the solution to a constrained optimization problem. If the set of clusters is $\{C_1,C_2, \ldots , C_n\}$ then we wish to select $I \subseteq \{1, 2, \ldots, n\}$ to maximize
\[
\sum_{i\in I} \sigma(C_i)
\]
subject to the constraint that, for all $i, j \in I$ with $i\neq j$, we have
\[
C_i\cap C_j = \emptyset
\]
That is, we wish to maximize the total persistence score over chosen clusters, subject to the constraint that clusters must not overlap.

We should note that while this explanation is the most compact of the three, it is also the least formal, and most heuristically motivated. 

\subsection{Topologically Motivated HDBSCAN*}\label{top}

Topological data analysis \cite{carlsson2009topology}, \cite{wasserman2016topology}, \cite{zomorodian2012topological} is a suite of techniques bringing the powerful tool of topology to bear on data analysis problems.  Recently the techniques of topological data analysis have been brought to bear on clustering problems \cite{carlsson2010multiparameter}, \cite{carlsson2013classifying}, \cite{chazal2014robust}, \cite{rinaldo2012stability},  \cite{rinaldo2010generalized}.   A number of insights can be gained by looking at HDBSCAN* through the lens of topological data analysis.  

The primary technique employed in topological data analysis is persistent homology \cite{edelsbrunner2008persistent}, \cite{zomorodian2005computing}. Although our description of HDBSCAN* makes use of persistent homology techniques, the full details of that subject are beyond the scope of this paper.  Please see \cite{carlsson2009topology} and \cite{ghrist2014elementary} for a good introduction to the topic. We will also  make use of the language of sheaves. Again, details are beyond the scope of this paper; see Ghrist \cite{ghrist2014elementary}, Mac Lane and Moerdijk \cite{maclane2012sheaves} or Bredon \cite{bredon2012sheaf} for an introduction to sheaves.

Persistent homology for analysis of ``point cloud data'' begins with the assumption we are presented with a set of data points $X$ living in a metric space $(\mathcal{X}, d)$. One can then construct a simplicial complex from this data. The standard approaches to this are the \emph{\v{C}ech complex} construction and the \emph{Vietoris-Rips complex} construction. Both approaches make use of a scale parameter $\varepsilon$ and open balls of radius $\varepsilon$ centered at a point $X_i$, denoted $B(X_i, \varepsilon) = \{x\in \mathcal{X} \mid d(X_i, x) < \varepsilon\}$. The \v{C}ech complex $C_\varepsilon$ is constructed by taking all elements of $X$ as 0-dimensional simplices, and adding an $n$-dimensional simplex spanning $X_{i_1}, X_{i_2}, \ldots X_{i_n}$ if the intersection $B(X_{i_1}, \varepsilon) \cap B(X_{i_2}, \varepsilon) \cap \cdots \cap B(X_{i_n}, \varepsilon)$ is non-empty. While the \v{C}ech complex has nice topological properties, it is often too computationally expensive to work with directly. The Vietoris-Rips complex $V_\varepsilon$ is constructed by, again, taking all elements of $X$ as 0-dimensional simplices, but adding an $n$-dimensional simplex spanning $X_{i_1}, X_{i_2}, \ldots X_{i_n}$ if all \emph{pairwise} intersections $B(X_{i_j}, \varepsilon) \cap B(X_{i_k}, \varepsilon)$ for $1\leq j \leq k \leq n$ are non-empty.

A family of simplicial complexes that have a natural ``nested" structure, such as $\{V_{\epsilon}\}_{\epsilon \geq 0}$, are called filtered simplicial complexes. There is a natural homology theory on filtered simplicial complexes, called persistent homology \cite{edelsbrunner2008persistent} \cite{zomorodian2005computing}. If we consider the 0-th homology, which computes groups with rank equal to the number of connected components of a topological space, we see that the persistent homology of the Vietoris-Rips complex associated to a point cloud provides a computation very similar to that of single-linkage clustering. 

As in the case of Robust Single Linkage we seek to make such a computation more robust to noise. Ultimately this falls to the method of construction of a simplicial complex from the point cloud data, and the metric of the space in which it resides. The goal is to make use of information about density in this construction. Intuitively, for both Vietoris-Rips and \v{C}ech complexes, higher dimensional simplices occur in denser regions of the space. Thus, the natural approach is to start with the Vietoris-Rips complex $V_\varepsilon$ and then remove all simplices that are not faces of simplices of dimension $k$. This gives a two parameter complex $W_{\varepsilon, k}$ where $k$ provides a density threshold. Unfortunately this approach, much like the \v{C}ech complex, is too computationally expensive to construct for all but trivial cases. Other alternative, but similar, approaches are proposed in \cite{martinez2012density} and \cite{martinez2016properties}, however we will follow Lesnick and Wright \cite{lesnick2015interactive} for a computationally tractable density-sensitive simplicial complex construction.

We begin with some notation. Given a simplicial complex $A$, define the $n$-skeleton of $A$, denoted $sk_n(A)$, to be the sub-complex of $A$ containing all simplices of $A$ of dimension less than or equal to $n$. Thus the 1-skeleton of a complex can be viewed as a graph, and the 0-skeleton as a discrete set of points. We define the Lesnick complex $L_{\varepsilon, k}$ as follows. Let $V_\varepsilon$ be the Vietoris-Rips complex associated to $X$ and define a graph $G_{\varepsilon, k}$ to be the subgraph of the 1-skeleton $sk_1(V_\varepsilon)$ induced by the vertices with degree at least $k$. The Lesnick complex $L_{\varepsilon, k}$ of $X$ is the maximal simplicial complex having 1-skeleton $G_{\varepsilon, k}$. For a fixed choice of $k$ we now have a filtered simplicial complex based on the family of complexes $\{L_{\varepsilon, k}\}_{\varepsilon > 0}$, and so we can apply standard persistent homology. 

To extend this topological approach to the full HDBSCAN* algorithm however we will take a slightly different approach to that described above. We draw upon the same fundamental ideas and intuitions, but use the language of sheaves. Intuitively a sheaf is a set that ``varies continuously'' over a topological space; thus each open set of the topological space has a set of \emph{sections} that lie above it. See \cite{ghrist2014elementary} or \cite{maclane2012sheaves} for further details.

Consider the set of non-negative reals $\mathbb{R}_{\geq 0} = \{x\in\mathbb{R} \mid x \geq 0\}$ with the following topology: for each non-negative real $x$ define an open set $\mathring{x} = \{y\in\mathbb{R}_{\geq 0} \mid y \geq x\}$ and take the topology formed by all such sets. Now define a sheaf $\mathscr{F}$ over $\mathbb{R}_{\geq 0}$ by defining

\begin{equation}
\mathscr{F}(\mathring{x}) = \{ sk_0(C) : C \in \Pi_0(L_{x, k}) \} ,
\end{equation}

where $\Pi_0(L_{x, k})$ is the set of connected components of the simplicial complex. That is, for an open set associated to the non-negative real $x$ we associate the set of 0-skeletons of connected components of the Lesnick complex $L_{x, k}$ associated to $X$. Since $\Pi_0$ is a functor, the maps from the filtered complex $L_{x, k} \hookrightarrow L_{y, k}$ for $x \leq y$ naturally induce restriction maps $res_{x,y}:\mathscr{F}(\mathring{x}) \to \mathscr{F}(\mathring{y})$. Verification that this is a sheaf is straightforward given the nested nature of the topology.

The sheaf is the structure we will use to capture persistence information; it can be seen as similar to a tree of clusters. While the specific sheaf described here is equivalent to a tree, the sheaf formalism allows the description of similar structures that cannot be described by trees. We now wish to condense the sheaf with regard to a parameter $m$ denoting the minimum cluster size. To do this we first consider the subsheaf $\mathscr{G}$ defined by

\begin{equation}
\mathscr{G}(\mathring{x}) = \{ s \in \mathscr{F}(\mathring{x})\: \mid\: |s| \geq m \}.
\end{equation}
This definition creates a simpler object by removing any sections that contain fewer than $m$ data points. Next we need to identify clusters from the sheaf. Since clusters must persist over a range distance scales (the $x$ in $L_{x,k}$) we must identify sections from different open sets. This can be viewed as the construction of an equivalence relation across the set of all sections in the sheaf
\[
S = \bigcup_{x\in \mathbb{R}_{\geq 0}} \mathscr{G}(\mathring{x}) .
\]
We define the equivalence relation as follows: given sections $s\in \mathscr{G}(\mathring{x})$ and $s'\in \mathscr{G}(\mathring{y})$ where (without loss of generality) $x \leq y$, we say $s$ is equivalent to $s'$ (denoted as $s\sim s'$) if and only if $res_{x,y}^{-1}(s') = \{s\}$ and for all $z$ with $x \leq z \leq y$ we have $|res_{z,y}^{-1}(s')| = 1$. That is, we consider sections at different distance scales equivalent if the section at the smaller distance scale is the \emph{only} section that restricts to the section at the larger distance scale, and this remains true for all intervening distance scales.

A cluster can then be identified with an equivalence class of sections under this equivalence relation. Such clusters necessarily overlap on the data points which they cover. If we wish to obtain a flat clustering we need to be able to score and compare clusters. We can score clusters in terms of their persistence over distance scales. Let $[s]$ be a cluster, and let $s_t \in S$ be the element of $S$ in the equivalence class $[s]$ that lies in $\mathscr{G}(\mathring{t})$, or the empty set if there is no representative of the equivalence class $[s]$ in $\mathscr{G}(\mathring{t})$. Define a function $\hat{s}(t) = |s_t|$, and then define the persistence score $\sigma$ of $[s]$ to be
\begin{equation}
    \sigma([s]) = \int_0^\infty \frac{\hat{s}(t)}{t^2} dt .
\end{equation}
The inclusion of the $\frac{1}{t^2}$ term provides the equivalent transformation to the shift from $\varepsilon$ to $\lambda = \frac{1}{\varepsilon}$ and ensures that this definition of $\sigma$ computes the same values as the definitions given in sections \ref{stats} and \ref{comp}.

To compare clusters for overlap it is necessary to be able to talk about the data points `in' a cluster. While the set of data points that make up a section is well defined, a cluster formed as an equivalence class of sections over different open sets has no natural assignment of data points to it. Instead we will define the points of a cluster $[s]$ to be the union of points in the sections within the equivalence class; thus we have a `points' function $p$ acting on clusters as 
\[
p([s]) = \bigcup_{t=0}^\infty s_t .
\]
The optimal flat clustering can then be described as the solution to a constrained optimization problem. If the set of clusters is $\{[s_1], [s_2], \ldots , [s_n]\}$ then we wish to select $I \subseteq \{1, 2, \ldots, n\}$ to maximize
\[
\sum_{i\in I} \sigma([s_i])
\]
subject to the constraint that, for all $i, j \in I$ with $i\neq j$, we have
\[
p([s_i])\cap p([s_j]) = \emptyset
\]
That is, we wish to maximize the total persistence score over chosen clusters, subject to the constraint that clusters must not overlap.

This is a complete description of the HDBSCAN* algorithm in topological terms. 

One of the major advantages of viewing HDBSCAN* through this lens is that it allows for generalisations that were not previously possible to describe. For example one could consider a new algorithm computing persistence across both $\varepsilon$ and $k$ simultaneously via techniques of multidimensional persistent homology, and making use of the more general structure of sheaves instead of trees. Such an approach provides a concrete realization of the techniques initially described in \cite{carlsson2010multiparameter}.  This would provide a new clustering algorithm, \emph{Persistent Density Clustering \cite{healy2017pdc}}, that is nearly parameter free.

\section{Accelerating HDBSCAN*}\label{accel}

As described in \cite{campello2013density} and \cite{campello2015hierarchical} the HDBSCAN* algorithm on $N$ data points has $O(N^2)$ run-time. To be competitive with other high performance clustering algorithms a sub-quadratic run-time is required, with an $O(N\log N)$ run-time strongly preferred. The run-time analysis of HDBSCAN* in \cite{campello2015hierarchical} identified three steps having $O(N^2)$ time complexity: the computation of core-distances (and mutual reachability distances); the computation of a minimum spanning tree (MST) used for single linkage computation; and the tree condensing. We propose to improve each of these steps, and in so doing, approach an average case complexity that grows approximately proportionally to $N \log N$.

One of the most common techniques for asymptotic performance improvement in the face of pairwise statistical problems (in our case pairwise distance computations) are space tree algorithms \cite{ram2009linear}. Indeed, these techniques are the basis for the impressive asymptotic performance of the DBSCAN and Mean Shift clustering algorithms, and are even used to accelerate some versions of K-Means. These techniques can also be applied to HDBSCAN* whenever the input data is provided as points in some metric space.

The computation of core-distances is a query for the $k^\text{th}$ nearest neighbor of each point in the input data set. The use of space tree algorithms for efficient nearest neighbor computations is well established. In particular kd-trees \cite{bentley1975multidimensional} in euclidean space, and ball-trees \cite{omohundro1989five} or cover trees \cite{beygelzimer2006cover} for generic metric spaces, provide fast asymptotic performance for nearest neighbor computation. Strict asymptotic run-time bounds for such algorithms are often complicated by properties of the data set. For example, cover tree nearest neighbor computation is dependent upon the expansion constant of the data, and the performance of kd-trees and ball-trees are similarly dependent upon the data distribution. However, an all points nearest neighbor query algorithm for cover trees with ``linear'' run-time complexity $O(c^{16} N)$, where $c$ is the expansion constant for the cover tree, is presented by Ram et al. \cite{ram2009linear}.  Claims of empirical run-time complexity of approximately $O(N \log N)$ for kd-trees and ball-trees are also common. While explicitly stating a run-time complexity for the core-distance computation is difficult, we feel confident in stating that, except for carefully constructed pathological examples, we can achieve sub-quadratic complexity.

With core-distance computation improved, the next challenge is the efficient computation of single linkage clustering using mutual reachability distance. In \cite{campello2015hierarchical}, Campello et al. use Prim's algorithm \cite{prim1957shortest} to compute a minimum spanning tree of the complete graph with edges weighted by the mutual reachability distance. Campello et al. then sort the edges, and use that data to construct the single linkage tree. Such an approach is similar to the SLINK algorithm \cite{mullner2011modern}, \cite{sibson1973slink} which essentially uses a modified version of Prim's algorithm (that does not explicitly compute an MST). For the purposes of computing a MST, Prim's is among the fastest available algorithms, however it is targeted toward graphs where the number of edges is some small multiple of the number of vertices, rather than complete graphs with $O(|V|^2)$ edges. In particular, if we have extra information about the vertices of the graph, other algorithms such as Bor\r{u}vka's algorithm \cite{boruuvka1926jistem} become more appealing. This is because if vertices are points in some metric space and edge weights are distances, Bor\r{u}vka's algorithm resembles a series of repeated all points nearest neighbor queries.

In \cite{march2010fast} March et al. make use of this observation and describe the Dual-Tree  Bor\r{u}vka algorithm for computing minimum spanning trees of points in a metric space. Given points $X$ in $(\mathcal{X}, d)$, they provide an algorithm to compute a minimum spanning tree of the weighted complete graph with vertices $X$ and edges $(X_i, X_j)$ with weight $d(X_i, X_j)$, where $X_i, X_j \in X$. The algorithm makes explicit use of space trees to provide impressive asymptotic performance. In particular, if cover trees are used, March et al. prove a run-time complexity of $O(\max\{c^6, c_p^2, c_l^2\} c^{10} N \log N \alpha(N))$, where $c$, $c_p$, and $c_l$ are data dependent constants and $\alpha$ is the inverse Ackermann function \cite{ackermann1928hilbertschen}. Here we provide a (minor) adaptation of the algorithm to compute a MST of the mutual reachability distances, resulting in a computation with sub-quadratic complexity.

In describing the algorithm we follow the approach of Curtin et al. in \cite{curtin2013tree} where they provide a version of March's algorithm adapted to a generic space partitioning tree framework. We begin with the introduction of notation to allow for easier statements of required algorithms.

For our purposes, a \emph{space tree} on a data set $X \subset (\mathcal{X}, d)$ is a rooted tree with the following properties:
\begin{itemize}
    \item Each node holds a number of points (possibly zero), has a single parent and has some number of children (possibly zero);
    \item each $X_i \in X$ is contained in at least one node of the tree;
    \item each node of the tree has an associated convex subset of $(\mathcal{X}, d)$ that contains all the points in the node, and the convex subsets associated with all of its children.
\end{itemize}
Notationally we will use a number of short form conventions to make discussions of points, children, descendants, and distances between nodes more convenient. Again, following Curtin et al. we will use the following notation:
\begin{itemize}
    \item The set of child nodes of a node $\mathscr{N}_i$ will be denoted $\mathscr{C}(\mathscr{N}_i)$ or simply $\mathscr{C}_i$ if the context allows.
    \item The parent node of a node $\mathscr{N}_i$ will be denoted $\mathscr{U}(\mathscr{N}_i)$.
    \item The set of points held in a node $\mathscr{N}_i$ will be denoted $\mathscr{P}(\mathscr{N}_i)$ or simply $\mathscr{P}_i$ if the context allows.
    \item The convex subset of $(\mathcal{X}, d)$ associated to a node $\mathscr{N}_i$ will be denoted $\mathscr{S}(\mathscr{N}_i)$ or simply $\mathscr{S}_i$ if the context allows.
    \item The set of \emph{descendant nodes} of a node $\mathscr{N}_i$, denoted by $\mathscr{D}^n(\mathscr{N}_i)$ or $\mathscr{D}^n_i$, is the set of nodes $\mathscr{C}(\mathscr{N}_i)\cup\mathscr{C}(\mathscr{C}(\mathscr{N}_i))\cup \ldots$ .
    \item The set of \emph{descendant points} of a node $\mathscr{N}_i$, denoted $\mathscr{D}^p(\mathscr{N}_i)$ or $\mathscr{D}^p_i$, is the set of points $\{p \mid p\in \mathscr{D}^n(\mathscr{N}_i)\cup\mathscr{P}(\mathscr{N}_i)\}$.
    \item The \emph{minimum distance} between two nodes $\mathscr{N}_i$ and $\mathscr{N}_j$, denoted $d_{\text{min}}(\mathscr{N}_i, \mathscr{N}_j)$ is defined as $\min \{ d(p_i, p_j) \mid p_i \in \mathscr{D}_i^p, p_j \in \mathscr{D}^p_j\}$.
    \item The \emph{maximum child distance} of a node $\mathscr{N}_i$, denoted $\rho(\mathscr{N}_i)$ is maximum distance from the centroid of $\mathscr{S}(\mathscr{N}_i)$ to any point in $\mathscr{N}_i$.
    \item The \emph{maximum descendant distance} of a node $\mathscr{N}_i$, denoted $\lambda(\mathscr{N}_i)$ is the maximum distance from the centroid of $\mathscr{S}(\mathscr{N}_i)$ to any \emph{descendant point} of $\mathscr{N}_i$.
\end{itemize}
In general, the minimum distance between nodes can be bounded below statically without having to compute all the point to point distances. For example, in kd-trees we have $d_{\text{min}}(\mathscr{N}_i, \mathscr{N}_j)$ bounded below by the minimum distance between $\mathscr{S}_i$ and $\mathscr{S}_j$ which can be computed at the time of tree construction without computing any point to point distances. Other types of space trees offer similar methods to bound node distances. 

In \cite{curtin2013tree} Curtin et al. provide a generic algorithm from which specific dual tree algorithms can be constructed. This provides a simple breakdown of a dual tree algorithm into core constituent parts, which the authors of this paper found particularly helpful in understanding March's algorithm. We therefore work within the same general framework here.

Dual tree algorithms make use of two different space trees, a \emph{query tree} $\mathscr{T}_q$ and a \emph{reference tree} $\mathscr{T}_r$. Curtin et al. breaks dual tree algorithms into three components. The first component is a \emph{pruning dual tree traversal}. This is a method of traversing a query and reference tree pair, pruning branches along the way. At each stage of such a pruning traversal we apply two procedures: the first, called \textsc{Score}, determines whether a branch is to be pruned (and potentially prioritises child branches); the second, called \textsc{BaseCase}, performs some algorithm specific operation on the pair of nodes at that stage of the traversal.

A simple approach to a dual tree traversal is a depth first traversal with no prioritisation of child nodes to explore. Algorithm \ref{alg:traversal} describes such an approach. In practice, one may want a more finely tailored traversal algorithm, with concomitant complexity of description, but for our explanatory purposes, this simple traversal is sufficient.

\begin{algorithm}
\caption{Depth First Dual Tree Traversal}\label{alg:traversal}
\begin{algorithmic}[0]
\Procedure{DepthFirstTraversal}{$\mathscr{N}_q, \mathscr{N}_r$}
\If{\textsc{Score}($\mathscr{N}_q, \mathscr{N}_r$) = $\infty$}
    \State \Return
\EndIf
\ForAll{$p_q \in \mathscr{P}_q, p_r \in \mathscr{P}_r$}
    \State \textsc{BaseCase}($p_q, p_r$)
\EndFor

\ForAll{$\mathscr{N}_{qc} \in \mathscr{C}_q, \mathscr{N}_{qr} \in \mathscr{C}_r$}
    \State \textsc{DepthFirstTraversal}($\mathscr{N}_{qc}, \mathscr{N}_{rc}$)
\EndFor
\EndProcedure
\end{algorithmic}
\end{algorithm}

Given a traversal algorithm, the specifics of March's Dual Tree Bor\r{u}vka algorithm now falls to the \textsc{BaseCase} and \textsc{Score} procedures. To explicate these we begin by describing Bor\r{u}vka's original algorithm, and then explain how we reconstruct it within a dual tree framework.

The general idea for Bor\r{u}vka's algorithm (Algorithm \ref{alg:boruvka}) is to build a forest, adding minimum weight edges to connect trees in iterative rounds. Bor\r{u}vka's algorithm starts with a weighted graph $G$, and initializes a forest $T$ to have the vertices of $G$, and no edges. Each pass of Bor\r{u}vka's algorithm finds minimum weight edges that span distinct connected components of $T$, and then adds those edges to $T$. As the algorithm proceeds, $T$ has larger but fewer connected components. The algorithm terminates when the forest $T$ is a single connected component, and thus a tree.

\begin{algorithm}
\caption{Classical Bor\r{u}vka's algorithm}\label{alg:boruvka}
\begin{algorithmic}[0]
\Procedure{MST}{$G = (V, E)$} 
\State $T \gets (V, \emptyset)$ \Comment{Initialize a graph $T$ with vertices from $G$ and no edges}
\While{$T$ has more than one connected component}
    \ForAll{components $C$ of $T$}
        \State $S \gets \emptyset$
        \ForAll{vertices $v$ in $C$}
            \State $D \gets \{a \in E \mid a\text{ meets }v\text{ and is not wholly contained in }C\}$
            \State $e \gets $ minimum weight edge in $D$
            \State $S \gets S\cup\{e\}$
        \EndFor
        \State $e \gets $ minimum weight edge in $S$
        \State Add $e$ to the graph $T$
    \EndFor
\EndWhile
\EndProcedure
\end{algorithmic}
\end{algorithm}

To convert Bor\r{u}vka's algorithm to a dual tree algorithm employing the spatial nature of the data, we make use of the space trees to find the nearest neighbors in a different component of the current forest for each point in the dataset. We then compile this information together to update the forest, and then reapply the nearest neighbor search.

Notationally we are building a forest $F$ with connected components $F_i$. At initialization $F$ has no edges, and there are $N$ connected components. At each pass of the algorithm we will add edges to $F$ and update the list of connected components accordingly. 

To keep track of state during processing, a number of associative arrays are required. First we require a mapping from points to the connected component of $F$ in which they currently reside. We denote this $\mathcal{F}$ and define $\mathcal{F}(p)$ to be the component $F_i$ which contains the point $p$. During the tree traversal we keep track of the nearest candidate point for each component with an associative array $\mathcal{N}$ such that $\mathcal{N}(F_i)$ is the candidate point (not in component $F_i$) nearest to component $F_i$ found so far. To keep track of which point in the component $F_i$ is closest to the candidate point we use an associative array $\mathcal{P}$ such that $\mathcal{P}(F_i)$ is the point in component $F_i$ nearest to $\mathcal{N}(F_i)$. Finally we keep track of the distance to a nearest neighbor for each component through an associative array $\mathcal{D}$ such that $\mathcal{D}(F_i)$ is the distance between $\mathcal{N}(F_i)$ and $\mathcal{P}(F_i)$.

To perform passes of the algorithm we need to use a modified nearest neighbor approach that looks for the nearest neighbor in a different component. Since we are searching for the ``nearest neighbors'' of the reference points each time, the query tree and reference tree are the same. After such an all-points nearest neighbor style tree search we can collate the results found for $\mathcal{N}$, $\mathcal{P}$ and $\mathcal{D}$ and use that to update the forest, and the associative array $\mathcal{F}$. This allows us to reset $\mathcal{N}, \mathcal{P}$ and $\mathcal{D}$ and make another pass with the same nearest neighbor style search. Each pass reduces the number of connected components in $F$ until we have a minimal spanning tree.

With this is mind, the \textsc{BaseCase} (algorithm \ref{alg:base}) needs to find points in different components that have a shorter distance separating them than the current value stored for the component under consideration. If such a pair is found we update $\mathcal{N}$, $\mathcal{P}$ and $\mathcal{D}$ accordingly.

\begin{algorithm}
\caption{Bor\r{u}vka's algorithm base case}\label{alg:base}
\begin{algorithmic}[0]
\Procedure{BaseCase}{$p_q, p_r$}
\If{$p_q = p_r$}
    \State \Return
\EndIf

\If{$\mathcal{F}(p_q) \neq \mathcal{F}(p_r)$ \textbf{and} $d(p_q, p_r) < \mathcal{D}(\mathcal{F}(p_q))$}
    \State $\mathcal{D}(\mathcal{F}(p_q)) \gets d(p_q, p_r)$
    \State $\mathcal{N}(\mathcal{F}(p_q)) \gets p_r$
    \State $\mathcal{P}(\mathcal{F}(p_q)) \gets p_q$
\EndIf
\EndProcedure
\end{algorithmic}
\end{algorithm}

The benefit of the tree based approach is that we are able to prune branches from our tree search which we know will not yield useful results. Since our queries are closely related to nearest neighbor queries we can make use of similar bounding approaches. The simplest such bound will prune the node pair $(\mathscr{N}_q, \mathscr{N}_r)$ if and only if the minimal distance between the nodes $d_{\text{min}}(\mathscr{N}_q, \mathscr{N}_r)$ is greater than the maximum of the nearest neighbor distances found so far for any point in $\mathscr{D}^p_q$. That is, if the closest any point in the query node (or its descendants) can be to any point in the reference node (or its descendants) is greater than all the current query node nearest neighbors found, clearly we do not need to descend any further.

In practice, with care and use of the triangle inequality, better bounds can be derived. We refer the reader to \cite{curtin2013tree} for the derivation, but note that we can define a bound
\[
\begin{split}
B(\mathscr{N}_q) = \min\Big\{ & \max\{ \max_{p\in\mathscr{P}_q} D_p, \max_{\mathscr{N}_c \in \mathscr{C}_q} B(\mathscr{N}_c) \},\\
 & \min_{p\in \mathscr{P}_q} \big(D_p + \rho(\mathscr{N}_q) + \lambda(\mathscr{N}_q)\big),\\
 & \min_{\mathscr{N}_c\in\mathscr{C}_q} \bigg( B(\mathscr{N}_c) + 2\big(\lambda(\mathscr{N}_q) - \lambda(\mathscr{N}_c)\big)\bigg),\\
 & B(\mathscr{U}(\mathscr{N}_q)) \Big\},
\end{split}
\]
where $D_p$ is the distance to the nearest neighbor of $p$ found so far. Given this bound, if $d_{\text{min}}(\mathscr{N}_q, \mathscr{N}_r) \geq B(\mathscr{N}_q)$ then we can safely prune the pair $(\mathscr{N}_q, \mathscr{N}_r)$. In practice pruning will be done based on the pre-computed lower bound estimate for $d_{\text{min}}(\mathscr{N}_q, \mathscr{N}_r)$. Furthermore, as the bound is expressed recursively, we can cache previous computations and calculate $B(\mathscr{N}_q)$ efficiently.

We can improve our pruning further by using component membership to prune: if all the descendant points in the query and reference nodes are in the same component then we do not need to descend and check any of those points. Again, this is a computation that can be done recursively and cached. With that in mind we can define a \textsc{Score} function (Algorithm \ref{alg:score}) that prunes away unnecessary branches, resulting in far fewer distance computations being required. 

\begin{algorithm}
\caption{Bor\r{u}vka's algorithm scoring}\label{alg:score}
\begin{algorithmic}
\Procedure{Score}{$\mathscr{N}_q, \mathscr{N}_r$}
\If{$d_{\text{min}}(\mathscr{N}_q, \mathscr{N}_r) < B(\mathscr{N}_q)$}
\If{$\forall (p_q\in\mathscr{D}^p_q, p_r\in\mathscr{D}^p_r) : \mathcal{F}(p_q) = \mathcal{F}(p_r)$}
    \State \Return $\infty$
\EndIf
\State \Return $d_{\text{min}}(\mathscr{N}_q, \mathscr{N}_r)$
\EndIf
\State \Return $\infty$
\EndProcedure
\end{algorithmic}
\end{algorithm}

Combining all these pieces together provides us with an algorithm to compute a minimum spanning tree of the distance weighted complete graph of points in a metric space. In practice, we wish to compute a MST using mutual reachability distance, and want to compute as few distances as possible. We can do this by using precomputed core-distances as a filter on the pairs of points passed to \textsc{BaseCase}, and only compute distances (and mutual reachability distances) for a subset of points. Furthermore, by prioritising the order in which we perform our dual tree traversal we can construct tighter bounds $B$ sooner, and thus perform more tree pruning, resulting in even fewer distances being computed.

We can express this in a more detailed tree traversal algorithm. The traversal algorithm is assumed to have access to the associative arrays $\mathcal{F}$ and $\mathcal{D}$. We also introduce a new associative array $\mathcal{C}$ such that $\mathcal{C}(p)$ is the core-distance (i.e. distance to the $k^{\text{th}}$ nearest neighbor) for the point $p$. We can then expand out the loop over pairs of algorithm \ref{alg:traversal} into a pair of nested for loops over points in the query node, and points in the reference node. This allows us to check if the core-distance of a point exceeds the current best distance for the component the point lies in. If the core-distance is larger then the mutual reachability distance is necessarily also larger, and hence this point can be eliminated from consideration.

\begin{algorithm}[!ht]
\caption{Tailored dual tree traversal}\label{alg:traversal2}
\begin{algorithmic}
\Procedure{DualTreeTraversal}{$\mathscr{N}_q, \mathscr{N}_r$}
\If{\textsc{Score}($\mathscr{N}_q, \mathscr{N}_r$) = $\infty$}
    \State \Return
\EndIf

\ForAll{$p_q \in \mathscr{P}_q$}
    \If{$\mathcal{C}(p_q) < \mathcal{D}(\mathcal{F}(p_q))$}
        \ForAll{$p_r \in \mathscr{P}_r$}
            \If{$\mathcal{C}(p_r) < \mathcal{D}(F(p_q))$}
                \State \textsc{BaseCase}($p_q, p_r$)
            \EndIf
        \EndFor
    \EndIf
\EndFor

\State $L \gets \textsc{Sort}([(\mathscr{N}_{qc}, \mathscr{N}_{rc}) \mid \mathscr{N}_{qc} \in \mathscr{C}_q, \mathscr{N}_{qr} \in \mathscr{C}_r],\,\, d_\text{min}(\cdot, \cdot))$
\For{$(\mathscr{N}_{qc}, \mathscr{N}_{qr})$ \textbf{in} $L$}
    \State \textsc{DualTreeTraversal}($\mathscr{N}_{qc}, \mathscr{N}_{rc}$)
\EndFor
\EndProcedure
\end{algorithmic}
\end{algorithm}

We can also prioritise the tree descent based on, for example, the distance between nodes, descending to nodes that are closer together first such that bounds get updated earlier. Algorithm \ref{alg:traversal2} gives an example of such a tailored algorithm that takes advantage of core-distances and prioritises descent down the tree. In practice the exact traversal and descent strategy can be more carefully tuned according to the exact space tree used.

It only remains to adapt the \textsc{BaseCase} procedure presented in Algorithm \ref{alg:base} to use core-distances to compute $d_{\text{mreach}}$ and use it instead of $d$ (as seen in Algorithm \ref{alg:base2}) and we have a Dual Tree Bor\r{u}vka algorithm adapted to perform HDBSCAN*. Such an algorithm allows us to compute a minimum spanning tree of the mutual reachability distance weighted complete graph without having to compute all pairwise distances. This results in an asymptotically sub-quadratic MST computation. While the data dependent nature of complexity analysis for tree based algorithms makes it difficult to place an explicit bound on the run-time complexity, analyses such as March et al. \cite{march2010fast} suggest we can certainly approach $O(N \log N)$ asymptotic performance for many data sets.

\begin{algorithm}[!ht]
\caption{HDBSCAN* tailored Bor\r{u}vka's algorithm base case}\label{alg:base2}
\begin{algorithmic}[0]
\Procedure{BaseCase}{$p_q, p_r$}
\If{$p_q = p_r$}
    \State \Return
\EndIf

\If{$F(p_q) \neq F(p_r)$}
    \State $\text{dist} = \max\{d(p_q, p_r), \mathcal{C}(p_q), \mathcal{C}(p_r)\}$
    \If{$\text{dist} < \mathcal{D}(\mathcal{F}(p_q))$}
        \State $\mathcal{D}(\mathcal{F}(p_q)) \gets \text{dist}$
        \State $\mathcal{N}(\mathcal{F}(p_q)) \gets p_r$
        \State $\mathcal{P}(\mathcal{F}(p_q)) \gets p_q$
    \EndIf
\EndIf
\EndProcedure
\end{algorithmic}
\end{algorithm}

One notable feature of mutual reachability distance is that it can result in many equal distances. We can exploit this fact within our modified Dual Tree Bor\r{u}vka algorithm. After a tree traversal the algorithm updates the forest $F$ and then resets the bounds $B$. Since there are many equal distances we can run the tree search again with the same bounds $B$ and find new potential edges to add to $F$. Such a run is extremely efficient as it has very tight bounds and thus rapidly prunes branches. We can repeat these runs until no new edges are found, and only then reset the values for $B$, forcing the algorithm to make fast progress in the face of ties, and near ties. Unfortunately this breaks the guarantee that the algorithm will also find a minimal spanning tree. However, in practice the result is a close approximation of a minimal spanning tree. We trade off a small loss in accuracy for a significantly faster algorithm. Furthermore, the minor differences in MST get smoothed out in tree condensing and flat cluster extraction process, resulting in very small deviations in final cluster results. This trade-off of performance for accuracy is particularly relevant for higher dimensional data sets when using kd-trees or ball-trees.

Given a minimal spanning tree it is possible to generate a single linkage cluster tree. This proceeds in two stages. The first stage is to sort the edges of the MST by weight (in this case, the mutual reachability distance between the pair of points the edge spans). Such an operation can be performed in $O(N \log N)$ run-time. In the second stage we process the edges in order using a union-find data structure \cite{tarjan1975efficiency}. This allows us to build the single linkage tree, providing the cluster merges and weights at which they occur, by progressively merging points and clusters by increasing weight. Since an MST has $O(N)$ edges we can complete this in $O(N\alpha(N))$ (using union-rank and path compression in our union-find algorithm).

The next step is to process the single linkage tree into a condensed tree. We can do this in a single pass working from the root in a breadth first traversal, building an associative array mapping single linkage cluster identifiers to new condensed tree cluster identifiers. At each node we need only check on the sizes of the child nodes, update the associative array accordingly, and record any data points falling out of the cluster. Since the single linkage tree has $N \log N$ nodes, the condensed tree processing can be completed in $O(N \log N)$ run-time.

In summary, the overall asymptotic run-time performance of the algorithm is bounded by the core-distance and minimum spanning tree computation stages, both of which now have sub-quadratic performance, and can be expected to approach $O(N \log N)$ performance for many data sets. This represents a significant improvement in potential scaling performance for HDBSCAN* clustering.

To test these algorthmic improvements we have implemented our accelerated HDBSCAN* algorithm in Python \cite{McInnes2017}. Our Python implementation builds from, and conforms to, the scikit-learn \cite{scikit-learn} software, making use of the kd-tree and ball tree data structures provided. Making use of scikit-learn has enabled our implementation to support a wide variety of distance metrics, as well as the ability to fall back to fast $O(N^2)$ algorithms when provided with a (sparse) distance matrix rather than vector space data. In the following section we will make use of our accelerated HDBSCAN* implementation to compare scaling of run-time performance with data set size with classic HDBSCAN*, and with other popular clustering algorithms. 

\section{Performance Comparisons}\label{perf}

In this section we will analyse the performance of our accelerated HDBSCAN*.  For the purpose of this paper we will not be considering the quality of clustering results as that has been adequately covered in \cite{campello2013density} and \cite{campello2015hierarchical}.  Instead we will demonstrate the computational competitiveness of our accelerated HDBSCAN* against other existing high performance clustering algorithms.  We are mindful of the difficulties of run-time analyses \cite{kriegel2016black}. We therefore focus on scaling trends with data set size (and dimension), and speak to the comparability of algorithms rather than making claims of strict superiority. 

All our run-time benchmarking was performed on a Macbook Pro with a 3.1 GHz Intel Core i7 processor and 8GB of RAM. Furthermore the benchmarking was performed in Jupyter notebooks which we have made available at \url{https://github.com/lmcinnes/hdbscan_paper}. We encourage others to verify and extend these benchmarks.

\subsection{Comparisons with HDBSCAN* reference implementation} 

As a baseline we compare the performance of our Python HDBSCAN* implementation against the reference implementation in Java from the original authors. Given two very different implementations in different languages our focus is on demonstrating that overall scalability and asymptotic performance can be improved through the spatial indexing acceleration techniques described.

\begin{figure}[!htb]
    \centering
    \includegraphics[width=0.9\textwidth]{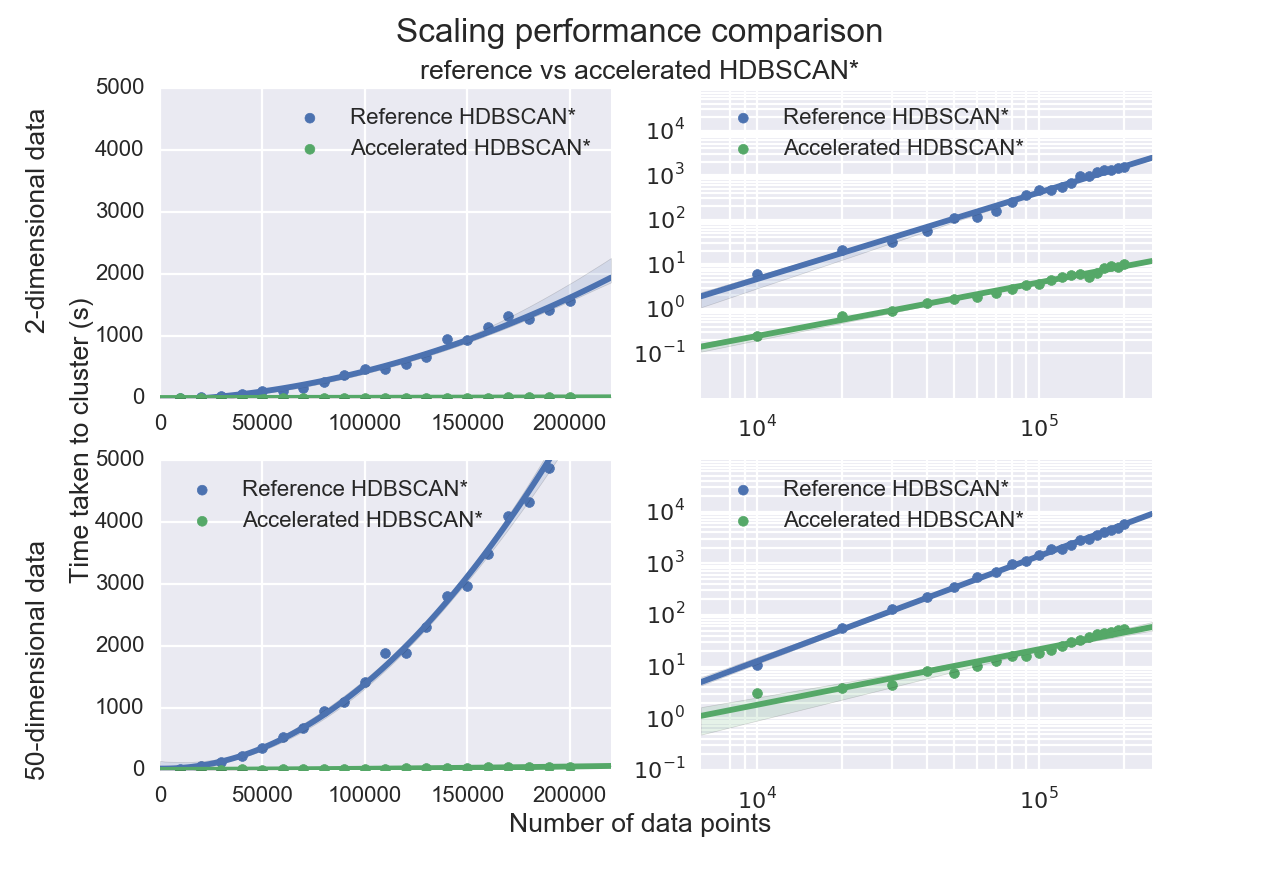}
    \caption{We compare the reference implementation in Java with the accelerated version implemented in Python. \protect\footnotemark}
    \label{fig:reference_impl_compare}
\end{figure}

We compare the performance on data sets of varying size for both 2-dimensional and 50-dimensional data. The results can be seen in Figure \ref{fig:reference_impl_compare}. The left hand column demonstrates raw performance times for both 2-dimensional and 50-dimensional data, while the right hand column provides a log-log plot that makes clear the different asymptotic performance of the algorithms. The accelerated Python version shows significantly improved performance, both in absolute terms, and asymptotically (having significantly lower linear slope in the log-log plot), clearly demonstrating sub $O(N^2)$ performance. Furthermore, in both the 2-dimensional and 50-dimensional cases, the accelerated Python version demonstrates roughly two orders of magnitude better absolute run-time performance on data set sizes of 200,000 points.

\footnotetext{See \url{https://github.com/lmcinnes/hdbscan_paper/blob/master/Performance\%20data\%20generation.ipynb} for the code used to generate this plot}

\subsection{Comparisons among clustering algorithms} 

In order to gain an overview of the performance landscape of clustering algorithms in general, we compare a number of the more popular clustering algorithms found in scikit-learn\footnote{Benchmarking was performed using scikit-learn v0.18.1.} \cite{scikit-learn} \cite{scikit-learn-github}. Since we recognise that implementation can have a significant effect on run-time performance, our goal here is merely to provide a sample of the performance space rather than direct comparisons to specific algorithms. We chose scikit-learn as it provides a number of techniques that all rest on a common implementation foundation (including our scikit-learn compatible HDBSCAN* implementation).

For the initial comparison we consider the following algorithms as implemented in scikit-learn: Affinity Propagation \cite{frey2007clustering}, Birch \cite{zhang1996birch}, Complete Linkage \cite{defays1977efficient}, DBSCAN \cite{ester1996dbscan}, KMeans \cite{macqueen1967kmeans}, Mean Shift\cite{fukunaga1975meanshift}, Spectral Clustering \cite{ng2001spectral}, and Ward Clustering \cite{ward1963hierarchical}. We compare these with our HDBSCAN* implementation.

Since this is a broad comparison of overall performance characteristics, each algorithm will be initialized with default scikit-learn parameters. In the next section, we do a more detailed comparison; carefully considering the impact of clustering algorithm parameters on performance.

\begin{figure}[!ht]
    \centering
    \includegraphics[width=0.8\textwidth]{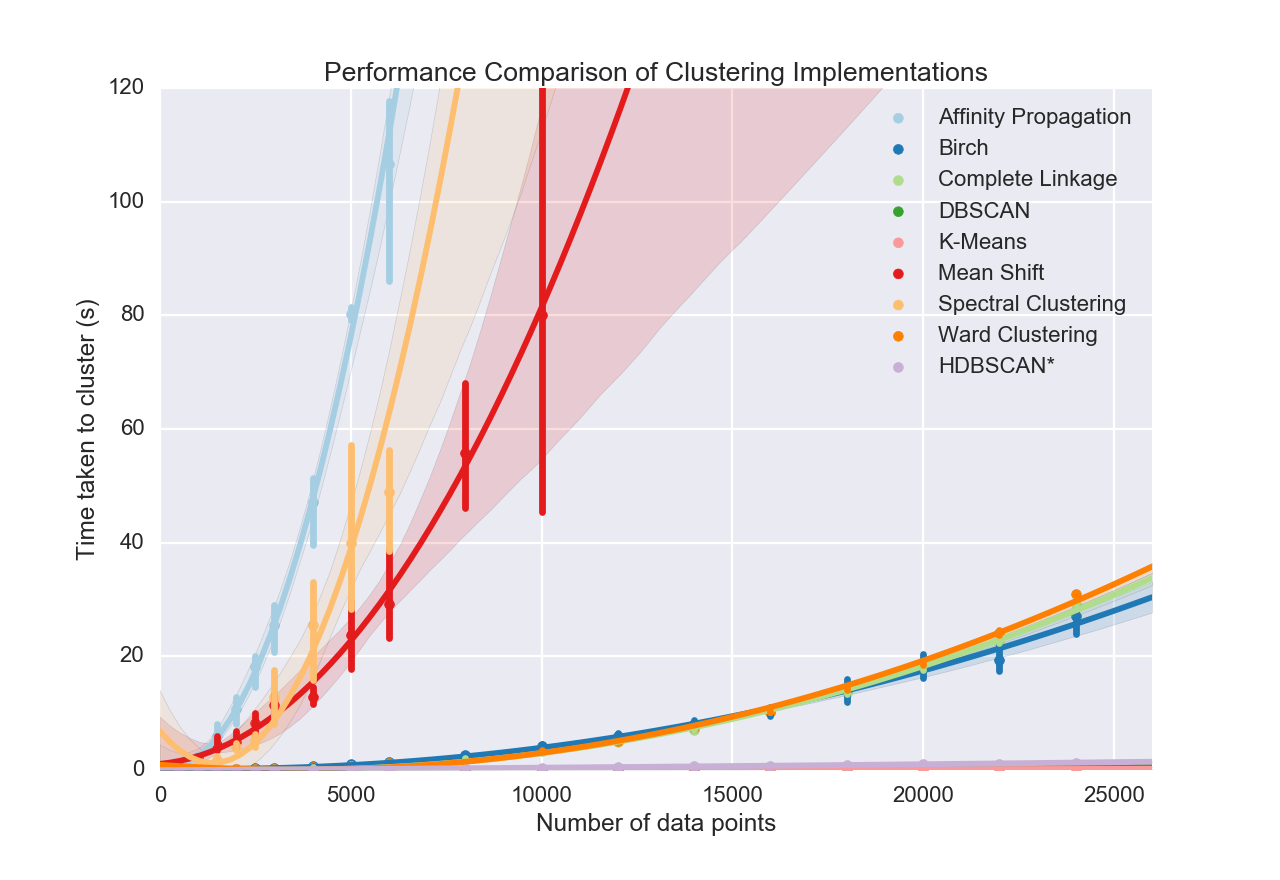}
    \caption{Comparison of scaling performance for scikit-learn implementations of a number of different clustering algorithms. Vertical bars present the range of run-times obtained over several runs at a given data set size.\protect\footnotemark}
    \label{fig:all_clustering_compare}
\end{figure}

\footnotetext{See \url{https://github.com/lmcinnes/hdbscan_paper/blob/master/Perfomance\%20comparisons\%20among\%20clustering\%20algorithms.ipynb} for the code used to generate this plot}

As demonstrated in Figure \ref{fig:all_clustering_compare}, there are three classes of implementation. The first is Affinity Propagation, Spectral Clustering, and Mean Shift, which all had poor performance beyond a few thousand data points. Some of this is undoubtedly implementation specific (particularly in the case of Spectral Clustering and Mean Shift). The next class of implementations are Ward, Complete Linkage and Birch, which performed better, but still scaled poorly for larger data set sizes. Finally, there was the group of DBSCAN, K-Means and HDBSCAN*, which are difficult to tell apart from one another in Figure \ref{fig:all_clustering_compare}.

\begin{figure}[!hbtp]
    \centering
    \includegraphics[width=0.8\textwidth]{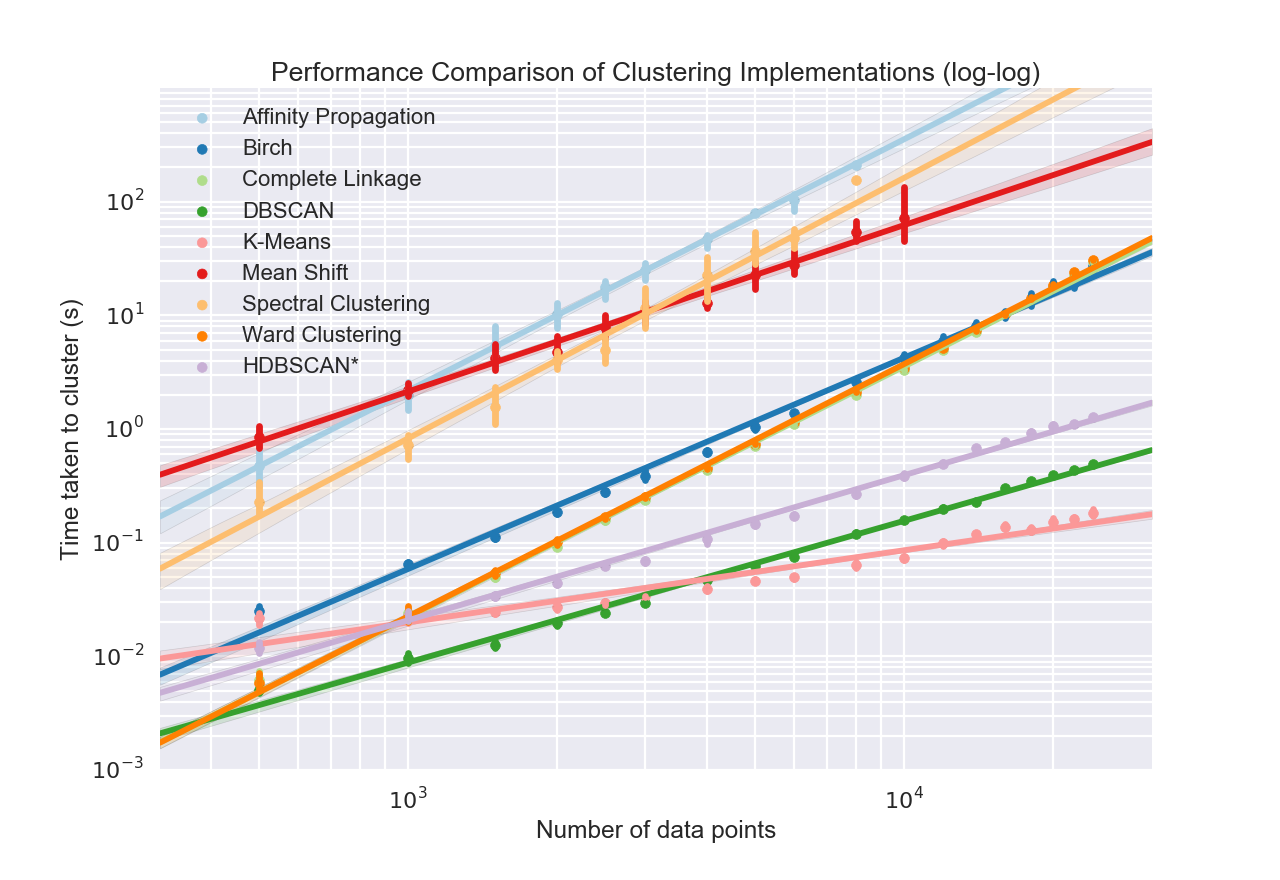}
    \caption{Comparison of scaling performance for scikit-learn implementations of a number of different clustering algorithms plotted on a log-log scale to demonstrate asymptotic performance more clearly.\protect\footnotemark}
    \label{fig:log_log_all_clustering_compare}
\end{figure}

\footnotetext{See \url{https://github.com/lmcinnes/hdbscan_paper/blob/master/Perfomance\%20comparisons\%20among\%20clustering\%20algorithms.ipynb} for the code used to generate this plot}

If we consider a log-log plot of the same data (Figure \ref{fig:log_log_all_clustering_compare}) in order to better see and understand the asymptotic scaling, we see algorithms ranging from K-Means impressive approximately $O(N)$ performance, through to the traditional $O(N^2)$ algorithms. DBSCAN and HDBSCAN* demonstrate similar asymptotics to each other, and are the closest in performance to K-Means. Also worth noting is that Mean Shift, while having poorer performance in general, has similar asymptotic performance to DBSCAN and HDBSCAN*.

K-Means, while being the fastest and most scalable algorithm (Figures \ref{fig:all_clustering_compare} and \ref{fig:log_log_all_clustering_compare}) explicitly fails to meet our desiderata. Although K-Means has only a single parameter, the selection of that parameter is difficult. K-Means also has implicit apriori assumptions about the data distribution -- specifically that clusters are Gaussian. Finally, K-Means is explicitly a partitioning algorithm and does not cope well with noise or outliers.

This leaves DBSCAN as the main competitor to our accelerated HDBSCAN*. We therefore seek a more detailed comparison of performance between DBSCAN and HDBSCAN*, specifically considering how parameter selection can affect performance.

\subsection{Comparisons with DBSCAN} 

The difficulty with DBSCAN run-time comparisons using default parameters is that very small values of $\varepsilon$ will return few or no core points. This results in a very fast run with virtually all the data being relegated to background noise.  Conversely, for large values of $\varepsilon$ DBSCAN will have very poor performance.  Our desire is to not misrepresent DBSCAN's run-times for real world use cases.  To circumvent this problem we will perform a search over the parameter space of DBSCAN in order to find the parameters which best match our HDBSCAN* results on a particular data set.  This is reasonable because, as described in section \ref{comp}, HDBSCAN* can be viewed as a natural extension to DBSCAN.  Once suitable parameters have been discovered we will benchmark the run-time of DBSCAN using those specific parameters against our HDBSCAN* run-time.  Of course, in practice a user may not, apriori, know the optimal parameter values for DBSCAN; that issue is not addressed in this experiment.  

As is the case for all tree based algorithms, run-time and run-time complexity are data dependent.  As indicated in \cite{kriegel2016black} this raises significant difficulties when benchmarking algorithms or implementations.  Our interest is in demonstrating the comparability of the scaling performance of these algorithms.   Under the assumption that both algorithms are tree based, they should have similar performance changes under different data distributions.  As such, we will examine the run-time behaviour of both algorithms with respect to a fairly simple data set.  One could extend this experimental framework to more complex data sets including those containing background noise.  

For this simple scaling experiment our data will consist of mixtures of Gaussian distributions laid down within a fixed diameter hypercube.  We use variable numbers of constant variance Gaussian balls for simplicity and to not unfairly penalize DBSCAN.  DBSCAN, as has been previously mentioned, does not support variable density clusters and thus could not match the output of HDBSCAN* in such cases.  We vary dimension, number of clusters and number of data points to determine their effect on run-time.

Although we are building a generative model on which to compare the performance of DBSCAN and HDBSCAN*, it should be noted that we are comparing the clustering results of the algorithms directly against each other and not against the underlying generative model.  This is intentional, since for any given instantiation of our generative model the generative model is not necessarily the most likely model (see \cite{Hennig2015clusters}).  We avoid this issue entirely by ignoring the generative model used to create the data for these experiments.

For the purposes of this experiment we chose to use scikit-learn's implementation of DBSCAN\footnote{Benchmarks were run using scikit-learn v0.18.1}. We did this for two reasons. First, scikit-learn's DBSCAN implementation is among the fastest of available DBSCAN implementations (see \cite{kriegel2016black} for DBSCAN implementation comparisons). Second, our Python HDBSCAN* implementation was built using scikit-learn\footnote{Our HDBSCAN* implementation has a similar level of genericity to DBSCAN, supporting the same distance metrics etc.}. This means that both the DBSCAN and HDBSCAN* implementations will be using the same underlying library implementations and, in particular, the same implementation of kd-trees which account for a significant part of any performance gains. A common implementation base aids in extension of results from implementations to algorithms.

\begin{figure}[!ht]
    \centering
    \includegraphics[width=1\textwidth]{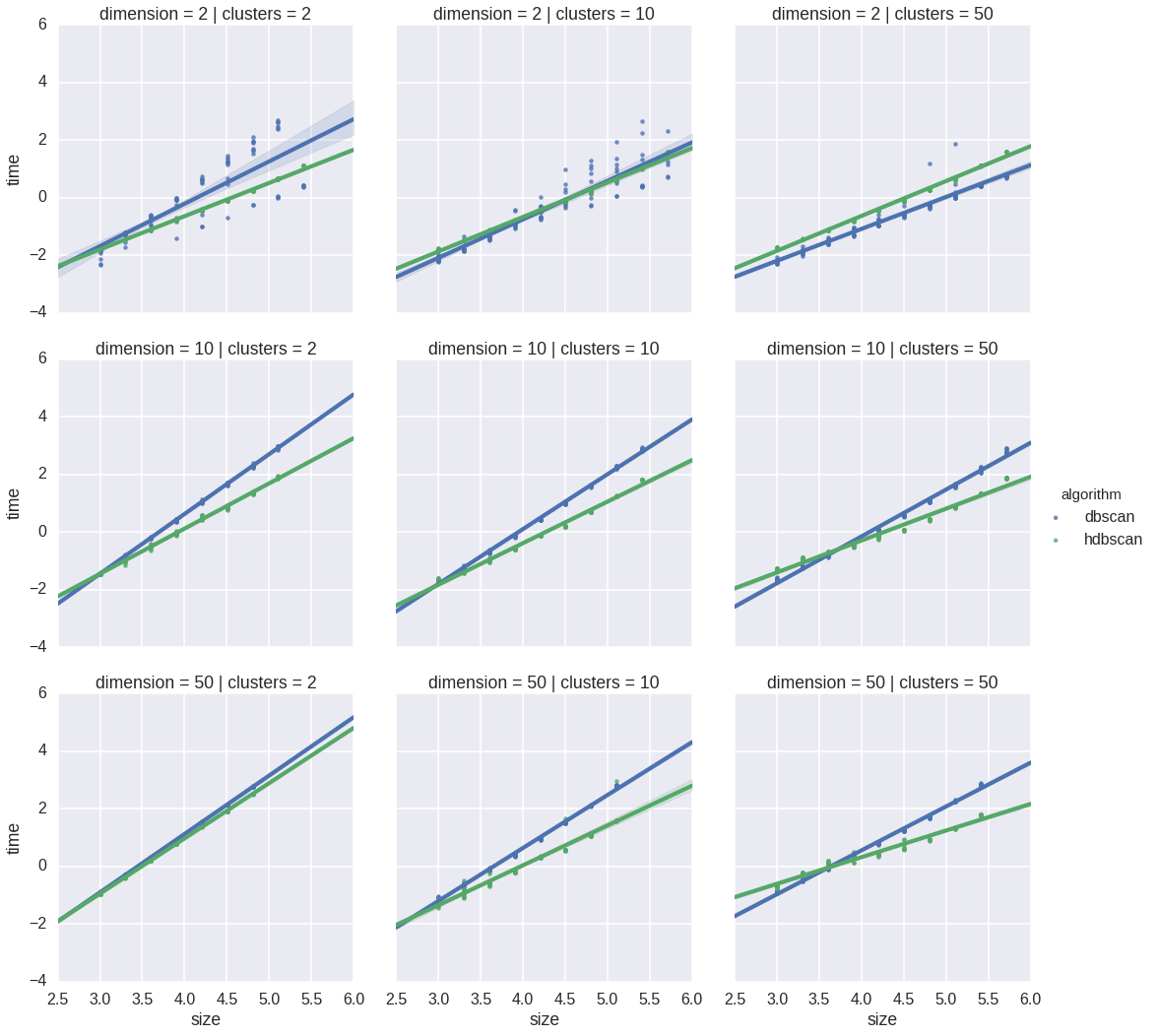}
    \caption{Comparison of scaling performance for scikit-learn's implementation of DBSCAN and our accelerated HDBSCAN*. Axes are on a $\log_{10}$ scale.  Each individual plot provides a log-log plot of run-time against data set size, with individual plots for each combination of data set dimension and number of clusters.  For each parameter combination ten random data sets where generated in order to assess the variation from data distribution.  The plot shows that accelerated HDBSCAN* and DBSCAN exhibit comparable performance.  \protect\footnotemark}
    \label{fig:dbscanVShdbscan}
\end{figure}

\footnotetext{A more detailed supplemental notebook can be found at \url{https://github.com/lmcinnes/hdbscan_paper/blob/master/Benchmark_vs_DBSCAN.ipynb}  }

In order to find the DBSCAN parameters of best fit to our HDBSCAN* clustering, we make use of the Gaussian process optimization framework within scikit-optimize \cite{scikit-optimize-github}.  We treat the background noise identified by both DBSCAN and HDBSCAN* as a single extra ``cluster'' which yields a partition, allowing us to use the adjusted Rand-index, as proposed by \cite{Hubert1985partitions}, to compute a similarity between the partitionings generated by each algorithm.  We then perform Gaussian process optimization to find the $\varepsilon$ and $k$ for DBSCAN which optimize this partition similarity score.  Due to the expense of this parameter search this optimization was distributed across multiple nodes of a large memory cluster\footnote{The results of this optimization can be found at \url{https://github.com/lmcinnes/hdbscan_paper/blob/master/optimizationResults.csv}. Due to the fact that we only care about relative timings between dbscan and hdbscan* we omit the exact specifications of this cluster}.  

The run-time comparison can be found in Figure \ref{fig:dbscanVShdbscan}. Each individual figure provides a log-log plot of run-time against data set size, with individual figures for each combination of data set dimension, and number of clusters.

It is worth noting that, particularly in the two dimensional case, due to crowding, the generative model can result in overlapped Gaussians and as a result HDBSCAN* produces a variable density clustering different from the constant density generative model. DBSCAN can have difficulty reproducing such a clustering. This leaves some open questions about the complete accuracy of the 2-dimensional run-times. However, the optimization process still often chose reasonable epsilon parameters so we feel that it is still representative of the broad expected performance for DBSCAN.

Figure \ref{fig:dbscanVShdbscan} clearly indicates that our accelerated algorithm has comparable asymptotic performance to DBSCAN.  Furthermore, our implementation has comparable absolute performance. This is a significant achievement considering that HDBSCAN* can be thought of as computing DBSCAN for all values of $\varepsilon$.  In fact, a single HDBSCAN* run allows a user to easily extract the DBSCAN clustering for any given $\varepsilon$.  More importantly, parameter selection and variable density clusters are DBSCANs challenges; our accelerated HDBSCAN* algorithm has overcome both of these challenges without sacrificing performance.  

\section{Future work}

A number of avenues for significant future work exist. First there are  several ways that our current Python implementation could be improved. The effects of approximate nearest neighbor search via spill trees \cite{liu2004investigation}, bounding adjustments \cite{curtin2015faster}, or RP-trees with local neighborhood exploration \cite{tang2016visualizing}, both on core-distance computation, and within March's algorithm remains unexplored. Approximate nearest neighbor computations may offer significant performance improvements for a small trade-off in the accuracy of results. Secondly, since there is no cover tree implementation for scikit-learn, our Python implementation does not support cover trees. Cover trees offer better scaling with ambient dimension (cover trees scale according to the expansion constant of the data, related to its intrinsic dimension), and support arbitrary distance metrics. A high performance cover tree implementation may provide significant benefits for our Python implementation of HDBSCAN*.

A significant weakness of our accelerated HDBSCAN* algorithm as described is that it is inherently serial. The inability to parallelise the algorithm is an obstacle for its use on large distributed data sets. We believe that partitioning the space via spill trees \cite{liu2004investigation}, and building MSTs on the partitioned data in parallel, then using the techniques of Karger, Klein and Tarjan \cite{karger1995randomized} to reconcile the overlapping trees may result in such a parallel algorithm. This is a topic of continued research.

Finally, the topological presentation of HDBSCAN* in section \ref{top} provides the opportunity to use multi-dimensional persistent homology to eliminate the parameter $k$. In such an approach no condensed tree interpretation is possible; instead the relevant structure is a sheaf over a partially ordered set (with the supremum topology). The resulting algorithm, \emph{Persistent Density Clustering}, is the subject of a forthcoming paper \cite{healy2017pdc}.

\section{Conclusions}

The HDBSCAN* clustering algorithm lies at the confluence of several threads of research from diverse fields. As a density based algorithm with a small number of intuitive parameters and few assumptions about data distribution, it is ideally suited to exploratory data analysis. In this paper we have described an accelerated HDBSCAN* algorithm that can provide comparable performance to the popular DBSCAN clustering algorithm. Since it has more intuitive parameters and can find variable density clusters, HDBSCAN* is clearly superior to DBSCAN from a qualitative clustering perspective. As the improvements of our accelerated HDBSCAN* make its computational scalability comparable in performance to DBSCAN, HDBSCAN* should be the default choice for clustering.

\bibliographystyle{acm}
\bibliography{main}

\end{document}